# Analysis of Dimensional Influence of Convolutional Neural Networks for Histopathological Cancer Classification

Alistair Michael Baretto
*Department of Computer Engineering*
Fr. Conceicao Rodrigues Institute of Technology, Vashi
Navi Mumbai, India
alistair.baretto@fcrit.onmicrosoft.com

Raj Sunil Salvi
*Department of Computer Engineering*
Fr. Conceicao Rodrigues Institute of Technology, Vashi
Navi Mumbai, India
raj.salvi@fcrit.onmicrosoft.com

Shreyas Rajesh Labhsetwar
*Department of Computer Engineering*
Fr. Conceicao Rodrigues Institute of Technology, Vashi
Navi Mumbai, India
shreyas.labhsetwar@fcrit.onmicrosoft.com

Piyush Arvind Kolte
*Department of Computer Engineering*
Fr. Conceicao Rodrigues Institute of Technology, Vashi
Navi Mumbai, India
piyush.kolte@fcrit.onmicrosoft.com

Veerasai Subramaniam Venkatesh
*Department of Computer Engineering*
Fr. Conceicao Rodrigues Institute of Technology, Vashi
Navi Mumbai, India
veerasai.subramaniam@fcrit.onmicrosoft.com

*Abstract—* Convolutional Neural Networks can be designed with different levels of complexity depending upon the task at hand. This paper analyzes the effect of dimensional changes to the CNN architecture on its performance on the task of Histopathological Cancer Classification. The research starts with a baseline 10-layer CNN model with (3 X 3) convolution filters. Thereafter, the baseline architecture is scaled in multiple dimensions including width, depth, resolution, and a combination of all of these. Width scaling involves inculcating a greater number of neurons per CNN layer, whereas depth scaling involves deepening the hierarchical layered structure. Resolution scaling is performed by increasing the dimensions of the input image, and compound scaling involve a hybrid combination of width, depth, and resolution scaling. The results indicate that histopathological cancer scans are very complex in nature and hence require high-resolution images fed to a large hierarchy of Convolution, MaxPooling, Dropout, and Batch Normalization layers to extract all the intricacies and perform perfect classification. Since compound scaling the baseline model ensures that all three dimensions: width, depth, and resolution are scaled, the best performance is obtained with compound scaling. This research shows that better performance of CNN models is achieved by compound scaling of the baseline model for the task of Histopathological Cancer Classification.

## I. Introduction

India has almost 32% of its population getting affected by cancer at some point in their lifetime. Cancer detection has always been an issue of major concern for the pathologists and medical practitioners for diagnosis and treatment planning. The manual detection of cancer through microscopic measures like biopsy images through histology is often subjective in nature and in most cases, varies from expert to expert.

Through the advancements of deep learning in the form of Convolutional Neural Networks, we can easily work on classification problems by detecting the core patterns in the dataset. CNNs have consistently been competitive with other techniques for image classification and recognition tasks. Nowadays, CNNs are outperforming their competing methodologies due to the availability of larger data sets, better models, and training algorithms and the availability of GPU (Cloud Computing) to enable the implementation of larger and deeper models. They have demonstrated excellent performance at tasks such as hand-written digit classification and facial detection. Several papers have further elucidated their capacity to deliver outstanding performance on more challenging visual classification tasks.

The objective of this work is to analyze the performance of CNN models subject to their dimensionality for the task of Histopathological Cancer Classification. In order to perform the task of histopathological cancer detection, the neural network must detect and learn the core differences between the two classes, namely, malignant and benign. A CNN architecture can be of different dimensions depending upon its purpose. For example, if an image classification task needs to be performed on images with few classes, the network shall have only a few layers and nodes, whereas on the other hand, if the problem was a multiclass classification problem, the architecture shall have relatively more number of layers and neurons per layer. Thus, this paper attempts to analyze and compare the classification performance of various CNN architectures, namely, baseline, depth scaled, width scaled, resolution scaled and compound scaled architecture for the binary classification task of histopathological scans into Malignant (Cancer Positive Metastatic Tissue) and Benign (Harmless) categories.

The paper is organised into multiple sections, the first section comprising the introduction to the topic. Second section highlights the related works, third section delves into the proposed work. Fourth section discusses the results of our research, and conclusion is presented in the fifth section.

## II. Related work

Mingxing Tan et al. [1] studied CNN model scaling and identified that optimal performance can be obtained by carefully balancing width, depth, and resolution scaling. They proposed a new scaling methodology that makes use of compound coefficient. They used this methodology to scale a baseline neural architecture to obtain a family of ConvNets titled as EfficientNets.

Or Sharir et al. [2] and Chenxi Liu et al. [3] studied the expressive powers of Neural Network Architectures, particularly, the effect of architectural similarity or overlapping upon the expressive efficiency between two architectures. They start their research using Convolutional Arithmetic Circuits (ConvACs) and extrapolate the results onto traditional ConvNets as well. They conclude that denser connectivity in the network architecture leads to an exponential increase in the expressive capacity of the neural networks.

Ningning Ma et al. [4] study the effect of speed, memory access cost, platform characteristics, and other direct factors upon the NN performance. They explore the effect of indirect metrics, such as FLOPs, and direct metrics upon the classification performance, and propose a new architecture called ShuffleNet V2 that exhibits exceptional speed and accuracy in image classification tasks.

Karen Simonyan et al. [5] and Dan Ciresan et al. [6] study the effect of depth scaling on the classification performance of ConvNets. Their research was particularly focused upon large scale image recognition using architectures with very small convolution filters (3 X 3), and the performance variations with depth scaling. They demonstrate that depth scaling had a positive impact on the classification accuracy for the ImageNet challenge dataset, and they generalize the results of depth scaling onto other classification tasks and complex recognition problems.

Andrew G. Howard [7] and Mathew D. Zeiler et al. [8] studied the effect of resolution scaling on the classification performance of CNNs. They explore the effect of inculcating image transformation in both the training dataset as well as the test dataset. They conclude that resolution scaling helped in improving the classification accuracy for the ImageNet Large Scale Visual Recognition Challenge.

Ali Sharif Razavian et al. [9] explore the generic descriptors extracted from CNNs, particularly from the OverFeat network to tackle a diverse range of object recognition tasks of image classification, scene recognition, fine-grained recognition, attribute detection, and image retrieval applied to a diverse set of datasets. They determine that, indeed, generic descriptors from complex CNNs are very powerful in object classification tasks.

Alex Krizhevsky et al. [10] analyze and perform the ImageNet classification task of 1.2 million high-resolution images into 1000 classes. They develop a custom 8-layered convolutional neural network consisting of 60 million parameters and 650,000 neurons. The CNN architecture comprises of 2 Convolutional Layers, 3 MaxPooling Layers, and 3 Fully-Connected (Dense) Layers. ReLu activation function is applied to the output of every convolutional and fully-connected layer. A final 1000-way softmax activation function is applied. Dropout regularization is applied to reduce overfitting in the fully-connected layers. The research demonstrates that the complexity of a CNN architecture depends primarily upon the complexity of the task at hand.

III. PROPOSED WORK

This research aims to design, implement, and compare the influence of dimensionality modifications in deep neural networks and suggest the best model for performing the task of histopathological cancer classification. For this purpose, the research involves training 5 different CNNs on the Kaggle dataset. The different networks would be based on the baseline network, a network with scaled width, scaled depth, scaled resolution, and a compound scaled network.

The presented work helps to understand the effect of networks with different dimensions on an image classification problem, in this case, the histopathological cancer classification. The initial analysis of the network design starts with a baseline architecture with few layers and having approximately optimal neural width. The subsequent networks are built on top of the baseline network by increasing the number of neurons per layer (width scaling), the addition of more layers (depth scaling), increasing the dataset image resolution (resolution scaling), and a combination of all these (compound scaling).

*A. Dataset Description*

The dataset used for this research is an open-source histopathological cancer detection dataset containing 160,000 images of which 144000 are used for training and 16000 for the purpose of validation [12,13]. The dataset consists of two classes viz, benign and malignant. The dataset is presented in the form of train and test folders with images and a separate CSV file containing the corresponding class labels for the images. A small sample of benign and malignant images of the dataset is shown in Fig. 1.

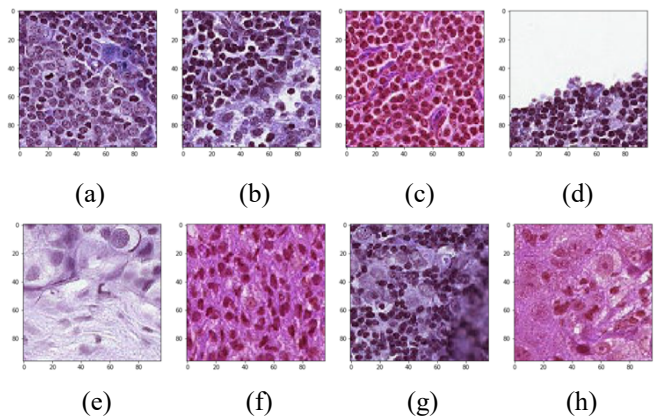

Fig. 1. (a,b,c,d) Benign Tissues, (e,f,g,h) Malignant Tissues

*B. Data Preprocessing*

The data is segregated into different directories depending upon its class label and the split it belongs to. Since the number of images is too large to train in a single batch for machine with lower RAM capacity, the research involves making use of generators to read images in batches with the help of ImageDataGenerator class of keras.preprocessing [14] as depicted in Fig. 2.

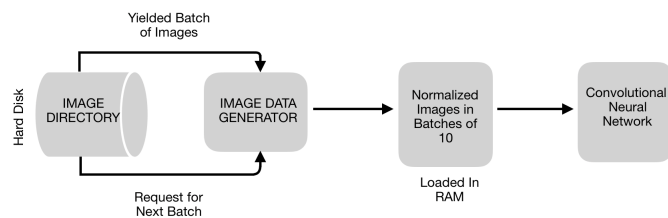

Fig. 2. Preprocessing Flow Diagram

Before this step, the data is normalized to have a uniform data distribution and help attain convergence faster while training.

*C. Convolutional Neural Network*

CNN is a Data Mining Algorithm that employs deep learning methodology for classification problems. It consists of a layered structure with three types of layers:

1. Convolution Layer:
2. Sub-sampling Layer (Max Pooling Layer)
3. Full Connection Layer

The complexity of a neural network depends upon the intricacy of the dataset involved as shown in Fig. 3, and can be scaled in multiple ways:

1. Width Scaling
2. Depth Scaling
3. Resolution Scaling
4. Compound Scaling

Width scaling involves increasing the number of neurons present in each NN layer. In the context of a CNN, it implies increasing the number of feature detectors in each Convolution and MaxPooling layer. Thus, the number of Feature Maps & Pooled Feature Maps increase in each layer. This helps the network extract and learn more intricate features per layer.

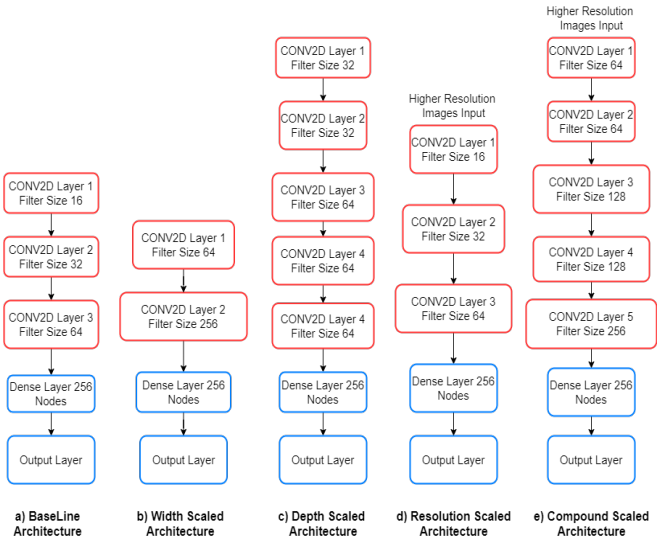

Fig. 3. CNN Architectures

Depth scaling involves the addition of more layers in the NN model. In the context of a CNN, it implies increasing the hierarchies of Convolution, MaxPooling, Dropout, and Batch Normalization layers. Although this helps the CNN model learn more complex features, depth scaling may result in the challenge of vanishing gradient descent, and training the network also becomes more difficult.

Resolution scaling involves increasing the number of pixels of the images on which the model is trained. This helps retain more information in the input images, thereby, increases the model's ability to extract and learn more details from the dataset. But, increasing the image resolution may be time-consuming, require higher RAM, and increase the time required for training the network.

Compound scaling involves increasing the CNN dimensions using one or more of the above methodologies. This research involves implementing a custom compound scaled model by increasing the width and depth of the CNN and also increasing the resolution of input images.

*D. Adam Optimizer*

Optimizers are algorithms that help compute the errors upon forward propagations and thus help in adjusting the attributes of a neural network such as its weights and learning rate in order to reduce the losses [15].

$$\omega_{t+1} = \omega_t - \frac{\alpha}{\sqrt{\hat{\lambda}_t} + \varepsilon} \cdot \hat{\mu}_t$$

Where,

$$\hat{\mu}_t = \frac{\mu_t}{1 - \gamma_1^t}$$

$$\hat{\lambda}_t = \frac{\lambda_t}{1 - \gamma_2^t}$$

$$\mu_t = \gamma_1 \cdot \mu_{t-1} + (1 - \gamma_1) \cdot \frac{\partial \Delta}{\partial \omega_t}$$

$$\lambda_t = \gamma_2 \cdot v_{t-1} + (1 - \gamma_2) \cdot \left[\frac{\partial \Delta}{\partial \omega_t}\right]^2$$

$\mu$ and $\lambda$ are initialized to 0
$\alpha$ is the Learning Rate
$\gamma_1 = 0.9$
$\gamma_2 = 0.999$
$\varepsilon$ is the Regularization Term

This research involves using Adam optimizer for all the networks in order to have a common base for training purposes. Adam Optimizer is a combination of momentum and RMSprop. It acts upon the gradient component using the exponential moving average of gradients (m) and the learning rate component by dividing the learning rate $\alpha$ by $\sqrt{v}$, the exponential moving average of squared gradients.
The initial learning rate is set to $\alpha = 0.0001$.

*E. Loss Function*

Training the Neural Networks involves an optimization process that employs a loss function to calculate the model error. A loss function in simple terms is an objective function that needs to be minimized. Loss functions can be broadly classified into two major categories depending upon the type of learning task, namely, Regression and Classification losses [16]. This research involves making use of a classification loss function called binary cross-entropy since the task at hand is a binary classification problem.

IV. RESULTS

The research involves training all the aforementioned CNN architectures with the open-source Histopathological Cancer Detection Dataset. The performance is measured using various performance metrics including Precision, Recall, F1-Score, and AUC of ROC curves. Also, performance graphs including Training and Validation Accuracy and Loss curves,

ROC curves, and Confusion Matrix are plotted for each CNN architecture trained for 20 epochs using Adam Optimizer.

### A. Baseline Model

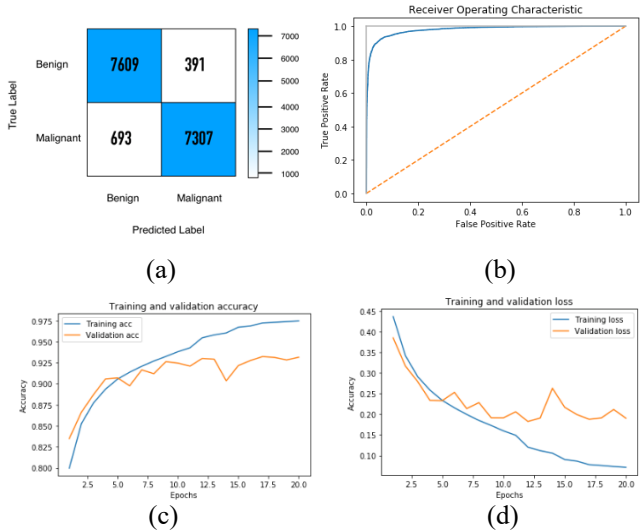

(a)        (b)

(c)        (d)

Fig. 4. (a) Confusion Matrix (b) ROC curve (c) Training and Validation Accuracy Curve (d) Training and Validation Loss Curve for Baseline Model

The Baseline Model successfully predicted 7609 benign and 7307 malignant out of 8000 and 8000 images respectively, while 391 benign got classified as malignant and 693 malignant got classified as benign. The AUC value for the ROC curve in Fig. 4(b) is 0.979. From Fig. 4(c), it is understood that the baseline model overfits the given dataset as there is a large deviation between training and validation accuracy after 10 epochs. Fig. 4(d) demonstrates that the training loss is decreasing and is about to flatten whereas the validation loss is having chaotic fluctuations. Hence, from all figures, we can conclude that the baseline model overfits the dataset. The baseline model's architecture is depicted in TABLE 1.

TABLE 1. Baseline CNN Architecture

| Layer | Output Shape | Parameters |
| --- | --- | --- |
| Conv2D | (None,106,106,16) | 448 |
| MaxPooling2D | (None,51,51,16) | 0 |
| Conv2D | (None,45,45,32) | 9248 |
| MaxPooling2D | (None,22,22,32) | 0 |
| Dropout | (None,22,22,32) | 0 |
| Conv2D | (None,2,2,128) | 147584 |
| MaxPooling2D | (None,1,1,128) | 0 |
| Flatten | (None,128) | 0 |
| Dense | (None,256) | 33024 |
| Dense | (None,2) | 514 |

### B. Width Scaling

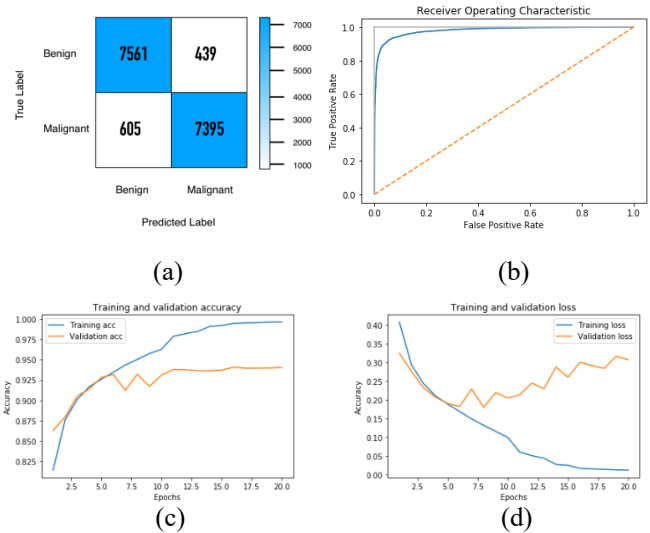

(a)        (b)

(c)        (d)

Fig. 5. (a) Confusion Matrix (b) ROC curve (c) Training and Validation Accuracy Curve (d) Training and Validation Loss Curve for Width Scaling Model

From Fig. 5(a), 7561 benign and 7395 malignant images are classified accurately out of 8000 images each, whereas 439 benign images got classified as malignant and 605 malignant images got classified as benign. The AUC value for the ROC curve in Fig. 5(b) is 0.983. The gap between Training and Validation accuracy as well as loss curve is increasing with the number of epochs as shown in Fig. 5(c) and Fig. 5(d). Thus, we conclude that width scaling is resulting in extreme overfitting.

### C. Depth Scaling

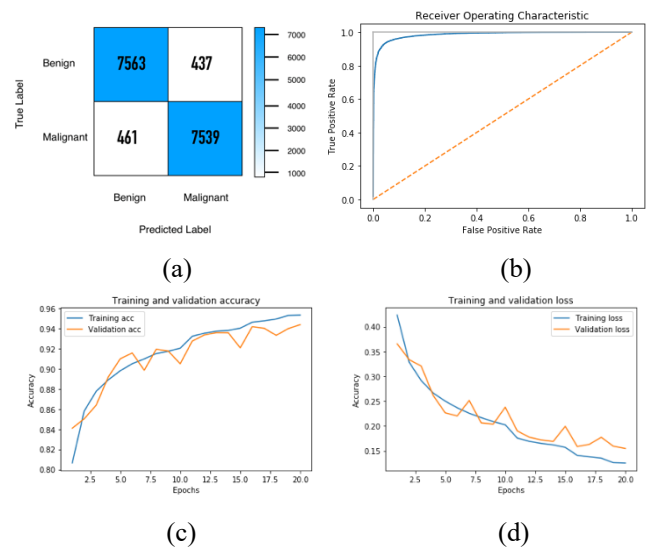

(a)        (b)

(c)        (d)

Fig. 6. (a) Confusion Matrix (b) ROC curve (c) Training and Validation Accuracy Curve (d) Training and Validation Loss Curve for Depth Scaling Model

From Fig. 6(a), 7563 benign and 7539 malignant images are classified accurately out of 8000 images each whereas 437 benign images got classified as malignant and 461 malignant images got classified as benign. The AUC value for the ROC curve in Fig. 6(b) is 0.985. Training accuracy and Validation accuracy is increasing and the gap between the curves is also small. Training loss and validation loss is decreasing. Thus, the Depth Scaling model does not overfit or underfit on the data but is also not considered as the most efficient model as validation accuracy and validation loss is fluctuating.

*D. Resolution Scaling*

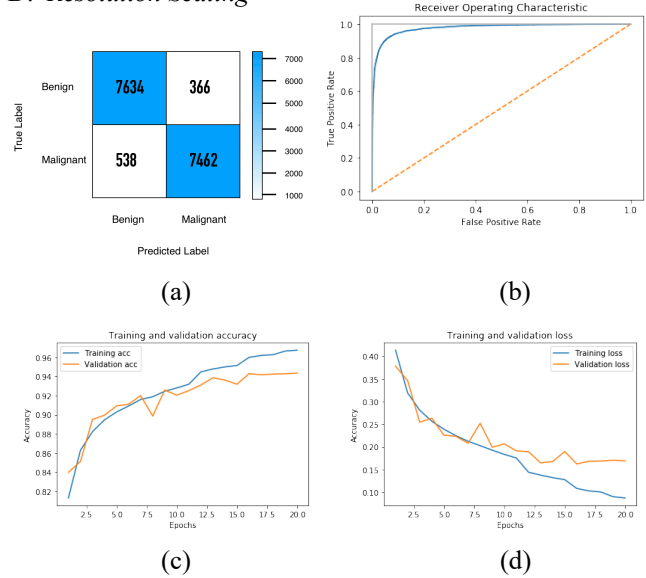

Fig. 7. (a) Confusion Matrix (b) ROC curve (c) Training and Validation Accuracy Curve (d) Training and Validation Loss Curve for Resolution Scaling Model

The Confusion Matrix in Fig. 7(a) shows that 7634 benign and 7462 malignant images are classified correctly, whereas 366 benign images are classified as malignant and 538 malignant images are classified as benign. The AUC value for the ROC curve in Fig. 7(b) is 0.983. From Fig. 7(c), it can be observed that the model works well up to the 7th epoch, after which the validation accuracy starts fluctuating and training accuracy keeps increasing resulting in overfitting.

*E. Compound Scaling*

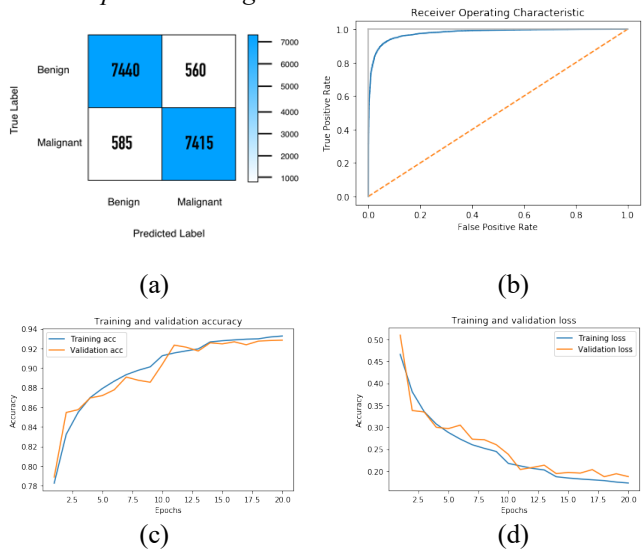

Fig. 8. (a) Confusion Matrix (b) ROC curve (c) Training and Validation Accuracy Curve (d) Training and Validation Loss Curve for Compound Scaling Model

The Confusion Matrix for Compound Scaling as shown in Fig. 8(a) demonstrates that 7440 benign and 7415 malignant images are correctly classified, whereas 560 benign images are classified as malignant and 585 malignant images are classified as benign. The AUC value for the ROC curve in Fig. 8(b) is 0.978. Graph from Fig. 8(c) shows that Training and Validation accuracy values are both increasing, and the gap between them is decreasing. Fig. 8(d) shows that both training and validation loss values are decreasing with epochs and are tending to flatten. Hence, from all the above observations, we conclude that Compound Scaling Model does not overfit or underfit on the data and is the most efficient model for the given dataset. The baseline model's architecture is depicted in TABLE 2.

TABLE 2. Compound Scaled CNN Architecture

| Layer | Output Shape | Parameters |
| --- | --- | --- |
| Conv2D | (None,106,106,16) | 448 |
| Conv2D | (None,104,104,16) | 2320 |
| Conv2D | (None,102,102,16) | 2320 |
| MaxPooling2D | (None,51,51,16) | 0 |
| Dropout | (None,51,51,16) | 0 |
| Conv2D | (None,49,49,32) | 4640 |
| Conv2D | (None,47,47,32) | 9248 |
| Conv2D | (None,45,45,32) | 9248 |
| MaxPooling2D | (None,22,22,32) | 0 |
| Dropout | (None,22,22,32) | 0 |
| Conv2D | (None,20,20,64) | 18496 |
| Conv2D | (None,18,18,64) | 36928 |
| Conv2D | (None,16,16,64) | 36928 |
| MaxPooling2D | (None,8,8,64) | 0 |
| Dropout | (None,8,8,64) | 0 |
| Conv2D | (None,6,6,128) | 73856 |
| Conv2D | (None,4,4,128) | 147584 |
| Conv2D | (None,2,2,128) | 147584 |
| MaxPooling2D | (None,1,1,128) | 0 |
| Dropout | (None,1,1,128) | 0 |
| Flatten | (None,128) | 0 |
| Dense | (None,256) | 33024 |
| Dropout | (None,256) | 0 |
| Dense | (None,2) | 514 |

It is understood from the above analysis that the performance of a CNN model varies with modifications in its architecture, as also evident in TABLE 3. Keeping the number of epochs constant ($\eta = 20$) and Adam optimizer ensures that these variations are primarily due to the architectural/hierarchical modifications.

TABLE 3. Performance Summary of CNN Architectures

| Classification Scores | Baseline Architecture | | Resolution Scaled Architecture | | Depth Scaled Architecture | | Width Scaled Architecture | | Compound Architecture | |
|---|---|---|---|---|---|---|---|---|---|---|
| | Benign | Malignant | Benign | Malignant | Benign | Malignant | Benign | Malignant | Benign | Malignant |
| Precision | 0.92 | 0.95 | 0.93 | 0.95 | 0.94 | 0.95 | 0.94 | 0.94 | 0.93 | 0.93 |
| Recall | 0.95 | 0.91 | 0.95 | 0.93 | 0.95 | 0.94 | 0.94 | 0.94 | 0.93 | 0.93 |
| F1-score | 0.93 | 0.93 | 0.94 | 0.94 | 0.94 | 0.94 | 0.94 | 0.94 | 0.93 | 0.93 |

## V. CONCLUSION

The impact of dimensional modifications to the performance of CNN models for the task of Histopathological Cancer classification is analyzed with the help of five architectures, namely baseline, width scaling, depth scaling, resolution scaling, and compound scaling. The classification performance is visualized by plotting Training and Validation Accuracy and Loss Curve, ROC curve, and Confusion Matrix for each CNN architecture. The research shows that the performance of CNN models improves with depth and resolution scaling of the baseline model. The histopathological cancer scans are very complex in nature, and hence require a deep hierarchy of Convolution, MaxPooling, Dropout, and Batch Normalization layers to extract all the intricacies and perform perfect classification. Resolution scaling ensures that all critical features persist in the training dataset. Since compound scaling of the baseline model ensures that all three dimensions: width, depth, and resolution are scaled, the best performance is obtained with compound scaling. Thus, the objective of analyzing the performance of CNN models and finding the best architecture for the task of Histopathological Cancer Classification is accomplished.

## ACKNOWLEDGMENT

We would like to thank our Professor Ms. M. Kiruthika, Department of Computer Engineering, Fr. C. Rodrigues Institute of Technology, Vashi for her support and guidance.